# An Incomplete Tensor Tucker decomposition based Traffic Speed Prediction Method

Jiajia Mi

*Abstract*—In intelligent transport systems, it is common and inevitable with missing data. While complete and valid traffic speed data is of great importance to intelligent transportation systems. A latent factorization-of-tensors (LFT) model is one of the most attractive approaches to solve missing traffic data recovery due to its well-scalability. A LFT model achieves optimization usually via a stochastic gradient descent (SGD) solver, however, the SGD-based LFT suffers from slow convergence. To deal with this issue, this work integrates the unique advantages of the proportional-integral-derivative (PID) controller into a Tucker decomposition based LFT model. It adopts two-fold ideas: a) adopting tucker decomposition to build a LFT model for achieving a better recovery accuracy. b) taking the adjusted instance error based on the PID control theory into the SGD solver to effectively improve convergence rate. Our experimental studies on two major city traffic road speed datasets show that the proposed model achieves significant efficiency gain and highly competitive prediction accuracy.

## I. INTRODUCTION

RECENTLY, traffic congestion has become increasingly serious due to explosive growth in the number of private cars. Intelligent transport systems (ITS) can effectively reduce traffic congestion [1], such as traffic route design. A vast amount of urban traffic data is collected and stored by a range of sensors to support traffic managers in their traffic decision-making, to provide real-time reference data for daily travel, as a result, could improve the functioning of the entire road network. However, we do not get complete and accurate spatiotemporal traffic data, but incomplete or incorrect data due to many non-human factors (e.g. sensor transmission distortion, detection equipment failure, extreme weather, etc.) [2]. For ITS, missing data has a negative impact on the analysis outcome, moreover, an extreme proportion of missing data will mislead conclusions for some spatiotemporal data analysis methods which demand high-integrity data and make them useless, thus the data loses its usability.

At present, there have been many proposed traffic data complementary models that exploit the temporal and spatial features of spatiotemporal data to enhance the complementary effect. For instance, Wang *et al.* [3] propose a model that incorporates matrix decomposition with simultaneous null consistency using a matrix approach. Qu *et al.* [4] propose an improved original PCA model to fit the distribution of the data based on probabilistic principal component analysis. Acar *et al.* [5] propose a deep learning model that utilized a graph convolutional bidirectional recurrent neural network architecture that learned intrinsic patterns based on historical data for data imputation. All the above methods have outstanding performance; however, they also demand high-quality traffic data for model training.

A latent factorization of tensors (LFT) based on Canonical/Polyadic factorization has a competitive performance in modeling and analyzing incomplete data [7-11]. Normally, researchers usually use stochastic gradient descent (SGD) to build a LFT-based learning model [12-15]. It is generally known that the SGD algorithm defines stochastic gradients based on instance errors which enable the LFT-based model obtains well generalizability. However, the SGD-based LFT model often suffers from low convergence rate [16-21].

A proportional-integral-derivative (PID) controller utilizes the past, current and future information of prediction error to control an automated system [22], which is commonly seen in industrial control applications, such as PID controllers for electric vehicle direct current motors [23-25], least squares gain PID controller for automatic driving [26], adaptive PID control method for self-reconfiguring systems [27]. Accordingly, as previous research [14-16], a PID-based optimization method can accelerate model convergence. Hence, this study adopts a tucker decomposition to build a LFT model and utilizes an SGD algorithm incorporating the PID principle as optimizer. This paper aims to make the following contributions:

1) Currently the LFT model is based on the Canonical/Polyadic decomposition framework, this paper will introduce the tucker decomposition framework, which obtains a larger feature representation space so as to construct an accurate tensor latent feature analysis method for traffic flow data.

2) We design an efficient tensor complementation scheme via incorporating the principle of PID to achieve high convergence rate.

Experimental results on two urban road traffic average speed datasets show that compared with advanced traffic data complementation methods, this model has a high prediction accuracy and still maintains a very competitive complementation accuracy even facing extreme missing ratios.

## II. PRELIMINARIES

### A. High-dimensional and incomplete speed tensor

The urban traffic speed data, i.e. a road segment × date × time tensor, is used as the basic input to perform urban traffic speed data analysis and extrapolation. As shown in Figure 1, the target tensor is a high-dimensional and incomplete (HDI) tensor, where the speed tensor contains a large number of unknown elements due to sensor failures or transmission problems. Therefore, our target tensor is defined as:

**Definition 1** (HDI road segment-date-time tensor): As for given three entity sets *I*, *J* and *K*, element of the tensor $\mathbf{Y}^{|I| \times |J| \times |K|}$, a road segment-date-time tensor, denotes the average speed of vehicles on road segment $i \in I$ at time $k \in K$ during day $j \in J$. Known sets $\Lambda$ in **Y** are significantly less than the unknown sets $\Gamma$, i.e. **Y** is HDI if $|\Lambda| \ll |\Gamma|$.

---

✧ J. J. Mi is with the School of Computer Science and Technology, Dongguan University of Technology, Dongguan, Guangdong 523808, China (e-mail: Mja_ii@163.com).

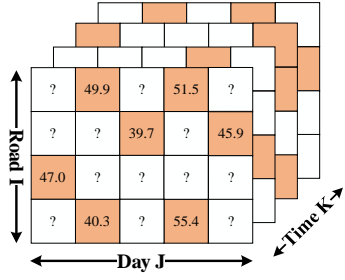

Fig 1. A tensor of road segment-date-time showing average speed.

*B. Latent factorization of tensors*

Note that the Canonical/Polyadic factorization (CPF) is commonly used to build the latent factorization of tensor (LFT), which has the advantages of simple model structure and low computational cost [23-26]. Compared to CPF, Tucker decomposition has a stronger multidimensional modelling capability. Due to its core tensor **G**, Tucker decomposition can obtain a larger feature representation space than CPT. Therefore, Tucker decomposition is used as the basic scheme for deriving this model in this paper.

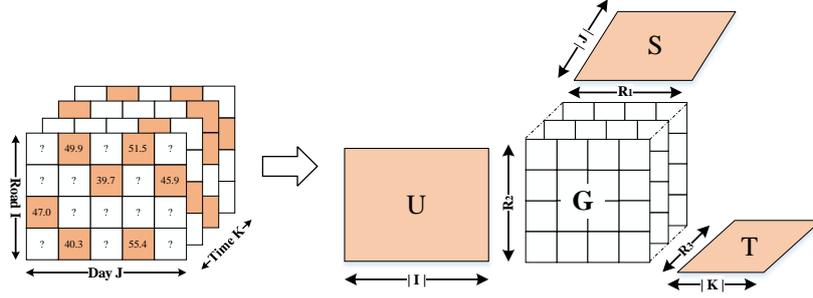

Fig 2. LF matrices and core tensor of LFs in TF-style.

***Definition* 2** (*Tucker decomposition*): The Tucker decomposition decomposes an original tensor into a core tensor and multiple factor matrices along each mode, as shown on Figure 2. The Tucker decomposition of a third-order tensor can be reformulated as follows:

$$\mathbf{Y} \approx \hat{\mathbf{Y}} = \mathbf{G} \times_1 \mathbf{U} \times_2 \mathbf{D} \times_3 \mathbf{T}. \quad (1)$$

where $\mathbf{U}^{|I| \times R_1}$, $\mathbf{D}^{|J| \times R_2}$, and $\mathbf{T}^{|K| \times R_3}$ denote three latent matrices, $\mathbf{G}^{R_1 \times R_2 \times R_3}$ denotes a core tensor. As a result, the element-wise expression of $\hat{\mathbf{Y}}$ can be obtained as:

$$\hat{y}_{ijk} = \sum_{m=1}^{R_1} \sum_{n=1}^{R_2} \sum_{l=1}^{R_3} g_{mnl} u_{im} d_{jn} t_{kl}. \quad (2)$$

We build an objective function based on the Euclidean distance to measure the difference between **Y** and $\hat{\mathbf{Y}}$ to obtain the desired U, D, T, and **G**. As is known, most of **Y** is the missing element. According to density-oriented modeling principles, the objective function is defined on known element set Λ [6]. Through it, we obtain the objective function as follows:

$$\varepsilon = \frac{1}{2} \sum_{y_{ijk} \in \Lambda} \left( y_{ijk} - \hat{y}_{ijk} \right)^2 = \frac{1}{2} \sum_{y_{ijk} \in \Lambda} \left( y_{ijk} - \sum_{m=1}^{R_1} \sum_{n=1}^{R_2} \sum_{l=1}^{R_3} g_{mnl} u_{im} d_{jn} t_{kl} \right)^2. \quad (3)$$

The known elements are not uniformly distributed in the HDI tensor **Y**, that is (3) is ill-posed. Thus, it is an outstanding approach to effectively prevent over-fitting of the model by introducing Tikhonov regularization [27]. With it, (3) is reformulated as follows:

$$\varepsilon = \frac{1}{2} \sum_{y_{ijk} \in \Lambda} \left( \left( y_{ijk} - \sum_{m=1}^{R_1} \sum_{n=1}^{R_2} \sum_{l=1}^{R_3} g_{mnl} u_{im} d_{jn} t_{kl} \right)^2 + \lambda_1 \sum_{m=1}^{R_1} \sum_{n=1}^{R_2} \sum_{l=1}^{R_3} g_{mnl}^2 + \lambda_2 \left( \sum_{m=1}^{R_1} u_{im}^2 + \sum_{n=1}^{R_2} d_{jn}^2 + \sum_{l=1}^{R_3} t_{kl}^2 \right) \right). \quad (4)$$

An SGD-based LFT has low computational effort and ease of implementation [31-33]. With it, the learning scheme is given as:

$$\arg\min_{\mathbf{U,D,T,G}} \varepsilon \xRightarrow{SGD} \forall i \in I, j \in J, k \in K, m,n,l \in \{1,2,...R\}:$$

$$u_{im}^{f+1} \leftarrow u_{im}^f - \eta \frac{\partial \varepsilon^f}{\partial u_{im}^f}, d_{jn}^{f+1} \leftarrow d_{jn}^f - \eta \frac{\partial \varepsilon^f}{\partial d_{jn}^f}, t_{kl}^{f+1} \leftarrow t_{kl}^f - \eta \frac{\partial \varepsilon^f}{\partial t_{kl}^f}, g_{mnl}^{f+1} \leftarrow g_{mnl}^f - \eta \frac{\partial \varepsilon^f}{\partial g_{mnl}^f}. \quad (5)$$

where $\varepsilon^f$ denotes the instantaneous loss on each training instance $y_{ijk} \in \Lambda$, $\eta$ is the learning rate of the SGD solver. Note that the stochastic gradient in (5) is given by:

$$\frac{\partial \varepsilon^f}{\partial u_{im}^f} = \lambda_2 u_{im}^f - e_{ijk}^f \sum_{n=1}^{R_2}\sum_{l=1}^{R_3} g_{mnl}^f d_{jn}^f t_{kl}^f, \quad \frac{\partial \varepsilon^f}{\partial d_{jn}^f} = \lambda_2 d_{jn}^f - e_{ijk}^f \sum_{m=1}^{R_1}\sum_{l=1}^{R_3} g_{mnl}^f u_{im}^f t_{kl}^f,$$
$$\frac{\partial \varepsilon^f}{\partial t_{kl}^f} = \lambda_2 t_{kl}^f - e_{ijk}^f \sum_{m=1}^{R_1}\sum_{n=1}^{R_2} g_{mnl}^f u_{im}^f d_{jn}^f, \quad \frac{\partial \varepsilon^f}{\partial g_{mnl}^f} = \lambda_1 g_{mnl}^f - e_{ijk}^f \sum_{m=1}^{R_1}\sum_{n=1}^{R_2}\sum_{l=1}^{R_3} u_{im}^f d_{jn}^f t_{kl}^f.$$

(6)

where $e_{ijk} = (y_{ijk} - \hat{y}_{ijk})$ denotes the instant error on $y_{ijk}$. By substituting (6) into (5), we obtain the parameters update rules.

$$\arg\min_{U,D,T,G} \varepsilon \overset{SGD}{\Rightarrow} \forall i \in I, j \in J, k \in K, m,n,l \in \{1,2,...R\}:$$

$$\begin{cases} u_{im}^{f+1} \leftarrow u_{im}^f - \eta\left(\lambda_2 u_{im}^f - e_{ijk}^f \sum_{n=1}^{R_2}\sum_{l=1}^{R_3} g_{mnl}^f d_{jn}^f t_{kl}^f\right), \\ d_{jn}^{f+1} \leftarrow d_{jn}^f - \eta\left(\lambda_2 d_{jn}^f - e_{ijk}^f \sum_{m=1}^{R_1}\sum_{l=1}^{R_3} g_{mnl}^f u_{im}^f t_{kl}^f\right), \\ t_{kl}^{f+1} \leftarrow t_{kl}^f - \eta\left(\lambda_2 t_{kl}^f - e_{ijk}^f \sum_{m=1}^{R_1}\sum_{n=1}^{R_2} g_{mnl}^f u_{im}^f d_{jn}^f\right), \\ g_{mnl}^{f+1} \leftarrow g_{mnl}^f - \eta\left(\lambda_1 g_{mnl}^f - e_{ijk}^f \sum_{m=1}^{R_1}\sum_{n=1}^{R_2}\sum_{l=1}^{R_3} u_{im}^f d_{jn}^f t_{kl}^f\right). \end{cases}$$

(7)

*C. A PID controller*

A PID controller adjust the observed error by proportional, integral and differential terms [29,30]. Figure 3 depicts the flow chart of a discrete PID controller. In this paper, we incorporate the PID control principle into an SGD-based learning scheme. Thus, given the true value (*TV*) and the predicted value (*PV*) at the *f* th time point, the discrete PID controller establishes the adjustment error as follows [14]:

$$\tilde{E}_f = K_P E_f + K_I \sum_{i=1}^{f} E_i + K_D (E_f - E_{f-1}).$$

(8)

where $E_f$ and $E_i$ denote the observation error at the *f*-th and *i*-th time points, as well as $K_P$, $K_I$ and $K_D$ denote the coefficients controlling the proportional, integral and derivative effects respectively. By taking a priori error into account, we obtain the adjusted error for achieving accurate control.

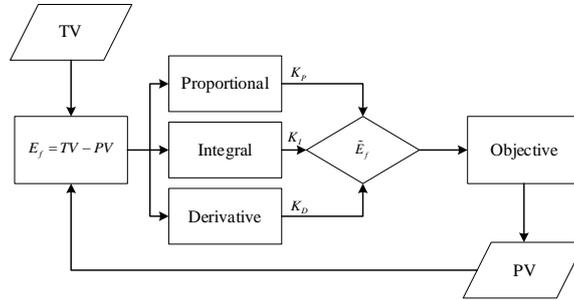

Fig 3. Flow chart of a discrete PID controller.

## III. OUR METHOD

Based on [6], introducing a linear bias into a learning model can improve the learning ability of the model. Considering speed tensor $\mathbf{Y}^{|I|\times|J|\times|K|}$, we introduce three linear bias vectors *a*, *b*, and *c* with length $|I|$, $|J|$, and $|K|$ into the learning model. Thus, we achieve the biased learning objectives as:

$$\begin{cases} \varepsilon = \frac{1}{2}\sum_{y_{ijk}\in\Lambda}\left((y_{ijk}-\hat{y}_{ijk})^2 + \lambda_1\sum_{m=1}^{R_1}\sum_{n=1}^{R_2}\sum_{l=1}^{R_3}g_{mnl}^2\right. \\ \left.+ \lambda_2\left(\sum_{m=1}^{R_1}u_{im}^2 + \sum_{n=1}^{R_2}d_{jn}^2 + \sum_{l=1}^{R_3}t_{kl}^2\right) + \lambda_3(a_i^2+b_j^2+c_k^2)\right) \\ \hat{y}_{ijk} = \mu + \sum_{m=1}^{R_1}\sum_{n=1}^{R_2}\sum_{l=1}^{R_3} g_{mnl}u_{im}d_{jn}t_{kl} + a_i + b_j + c_k. \end{cases}$$

(9)

where $\mu = \left(\sum_{y_{ijk}\in\Lambda} y_{ijk}\right)/|\Lambda|$ is applied as a global parameter to approximate the overall mean of the tensor elements.

As shown in (7), $e_{ijk}^{f} = \left(y_{ijk} - \hat{y}_{ijk}^{f}\right)$ is the instantaneous error in $f$ th iteration of the instance $y_{ijk}$ in SGD solver. It can be seen from (8) that we can consider the instantaneous error as a proportional term in the PID controller. Therefore, a simplified PID controller can be used to describe the above process by setting $K_P = 1$ and $K_I = K_D = 0$.

Hence, according to the PID control principle presented in (8) [14], the following complete expression for the adjusted instance error of $y_{ijk}$ at the $f$ th iteration is established:

$$\tilde{e}_{ijk}^{f} = K_P e_{ijk}^{f} + K_I \sum_{r=1}^{f} e_{ijk}^{r} + K_D \left(e_{ijk}^{f} - e_{ijk}^{f-1}\right). \tag{10}$$

Note that the effect of (10) can be interpreted from the aspect of "learning residual" by a learning algorithm:

a) The proportional term $e_{ijk}^{f}$: It can be denoted as the current learning residual of the SGD, and adjusted by proportional coefficient $K_P$;

b) The integral term $\sum_{r=1}^{f} e_{ijk}^{r}$: It can be denoted as the past learning residuals associated with the instance $y_{ijk}$ and adjusted by the $K_I$ integral coefficient. Such term regulates the model's learning direction to avoid oscillations;

c) The derivative term $\left(e_{ijk}^{f} - e_{ijk}^{f-1}\right)$: It can be expressed as the instant modification of the learning residuals and adjusted by a derivative coefficient $K_D$, which controls the model's future expectations to prevent model overshoot.

Thus, by substituting (10) into (9), we obtain a PID-incorporated learning scheme for desired learning parameters as:

$$\begin{cases} u_{im}^{f+1} \leftarrow u_{im}^{f} - \eta \left(\lambda u_{im}^{f} - \tilde{e}_{ijk}^{f} \sum_{n=1}^{R_2} \sum_{l=1}^{R_3} g_{mnl} d_{jn}^{f} t_{kl}^{f}\right), a_{i}^{f+1} \leftarrow a_{i}^{f} - \eta \left(\lambda a_{i}^{f} - \tilde{e}_{ijk}^{f}\right); \\ d_{jn}^{f+1} \leftarrow d_{jn}^{f} - \eta \left(\lambda d_{jn}^{f} - \tilde{e}_{ijk}^{f} \sum_{m=1}^{R_1} \sum_{l=1}^{R_3} g_{mnl} u_{im}^{f} t_{kl}^{f}\right), b_{j}^{f+1} \leftarrow b_{j}^{f} - \eta \left(\lambda b_{j}^{f} - \tilde{e}_{ijk}^{f}\right); \\ t_{kl}^{f+1} \leftarrow t_{kl}^{f} - \eta \left(\lambda t_{kl}^{f} - \tilde{e}_{ijk}^{f} \sum_{m=1}^{R_1} \sum_{n=1}^{R_2} g_{mnl} u_{im}^{f} d_{jn}^{f}\right), c_{k}^{f+1} \leftarrow c_{k}^{f} - \eta \left(\lambda c_{k}^{f} - \tilde{e}_{ijk}^{f}\right); \\ g_{mnl}^{f+1} \leftarrow g_{mnl}^{f} - \eta \left(\lambda g_{mnl}^{f} - \tilde{e}_{ijk}^{f} \sum_{m=1}^{R_1} \sum_{n=1}^{R_2} \sum_{l=1}^{R_3} u_{im}^{f} d_{jn}^{f} t_{kl}^{f}\right). \end{cases} \tag{11}$$

## IV. EXPERIMENTAL RESULTS AND DISCUSSION

### A. General Settings

*1) Evaluation Metric:* In this study, we focus on prediction accuracy and computational efficiency. Therefore, we use the Root Mean Squared Error (RMSE) as evaluation metric. If $\hat{y}_{ijk}$ and $y_{ijk}$ denote the estimated and actual values respectively, the expressions can be written in the following form:

$$RMSE = \sqrt{\sum_{y_{ijk} \in \Omega} \left(y_{ijk} - \hat{y}_{ijk}\right)^2 / |\Omega|}.$$

*2) Datasets:* We conduct an empirical study on two publicly available datasets collected from real traffic systems.

**D1:** The Shanghai Urban Road Speed Dataset1 is collected from 18 road sections in Shanghai, China, with a single daily time interval of five minutes (288-time intervals per day) over a period of 28 days.

**D2:** Seattle Urban Road Speed Dataset[34] collected from 323 loop detectors in Seattle, USA, with a single daily time interval of five minutes (288-time intervals per day) over a period of 28 days (i.e.2015.1.1-2015.1.28).

Aiming to obtain objective experimental results, we divided each dataset into training, validation, and test sets with the ratio of 8%:2%:90%. This means that we randomly select 8% of D1 as the training set, 2% as the validation set to build the model and tune the hyperparameters, and the remaining 90% of D1 will be used to evaluate its performance. In addition, the dimension of the LF space $R$ is set to 5 for a fair comparison, so as to eliminate the effect of hyperparameters on all models involved. And the procedure described above was repeated 20 times in order to eliminate possible biases arising from data segmentation. Attention is drawn to the fact that if the difference between the validation errors of two consecutive iterations is less than $10^{-5}$, or if the number of iterations exceeds a preset threshold of 1000, the model under test is judged to have converged.

### B. Compared Models

In this section, we compare our model with two state-of-the-art models. They are listed as follows:

**M1:** an iterative tensor decomposition method to solve the missing data problem [40]. It utilizes observations to correct missing data in order to improve model accuracy.

**M2:** It proposed a Gradient Descent based tensor evaluation model to construct multidimensional relationships between data for maintaining accuracy in an open and dynamic environment [38].

**M3:** our model in this work.

---

1. https://github.com/sysuits/High-dimensional-traffic-data-analysis

## C. Experimental Results

Tables II summarizes the RMSE and Total time cost of M1-M3. From the results, we can find **M3 has highly competitive prediction accuracy for missing data of an HDI tensor compared with its peers.** On D1, M3's RMSE is 10.4078, an accuracy improvement of 6.41% compared to M1's 11.1213, and 44.17% compared to M2's 18.6428, respectively. On D2, the RMSE for M3 is 7.0234, an accuracy improvement of 7.17% compared to M1's 7.5665, and 26.35% compared to M2's 9.5364, respectively. **Furthermore, M3 is strongly competitive with similar algorithms for computational efficiency**. On D1, the total time of M3 is 2 seconds to converge in RMSE, which is 95.40% of 37 seconds by M1, and 60% of 5 seconds by M2, respectively. Correspondingly, on D1 the total time cost of M5 spend in RMSE is 890 seconds to converge, which is 89.79% of 8782 seconds by M1, and 19.31% of 1103 seconds by M2, respectively.

TABLE II. Comparison of Each Model on All Testing Case.

| Testing Case | | M1 | M2 | M3 |
|---|---|---|---|---|
| **D1** | RMSE | 11.1213 | 18.6428 | **10.4078** |
| | Total time cost | 37 | 5 | **2** |
| **D2** | RMSE | 7.5665 | 9.5364 | **7.0234** |
| | Total time cost | 8782 | 1103 | **890** |

## V. CONCLUSION

For fast and accurate prediction of missing values in urban traffic road average speed data, a non-negative latent factorization tensor model based on Tucker decomposition is proposed, which considers the past, current, and future update information in parameter updates process. Validation results on two urban road traffic datasets show that it is highly competitive in terms of prediction accuracy. In addition, the design and implementation of adaptive algorithms for parameter tuning will be considered in the future [43-47].

REFERENCES

[1] Kai, Y. Y. "An introduction to intelligent transport system and the countermeasure of its development in china." *Progress in Geography* 18, no. 3 (1999): 274-278.

[2] H. Wu, X. Luo, and M. C. Zhou, "Discovering hidden pattern in large-scale dynamically weighted directed network via latent factorization of tensors," *in Proceedings of the 2021 IEEE 17th International Conference on Automation Science and Engineering (CASE)*, pp. 1533-1538, 2021.

[3] Y. Wang, Y. Zhang, X. Piao, H. Liu, and K. Zhang, "Traffic data reconstruction via adaptive spatial-temporal correlations," *IEEE Transactions on Intelligent Transportation Systems*, vol. 20, no. 4, pp. 1531–1543, 2019.

[4] L. Qu, J. M. Hu, L. Li, and Y. Zhang, "PPCA-based missing data imputation for traffic flow volume: A systematical approach," *IEEE Transactions on Intelligent Transportation Systems*, vol. 10, no. 3, pp. 512–522, 2009.

[5] E. Acar, D. M. Dunlavy, T. G. Kolda, and M. Mørup, "Scalable tensor factorizations with missing data," *in Proc. of the 2010 SIAM International Conference on Data Mining*, 2010.

[6] X. Luo, H. Wu, H. Yuan, and M. C. Zhou, "Temporal pattern-aware QoS prediction via biased non-negative latent factorization of tensors," *IEEE Transactions on Cybernetics*, vol. 50, no. 5, pp. 1798–1809, 2020.

[7] M. Z. Chen, H. Wu, C. He, and S. Chen "Momentum-incorporated latent factorization of tensors for extracting temporal patterns from QoS data," *in Proceedings of the 2019 IEEE International Conference on Systems, Man and Cybernetics (SMC)*, Bari, Italy, pp. 1757-1762, 2019.

[8] X. Luo, Y. Zhou, Z. Liu, and M. C. Zhou, "Fast and accurate non-negative latent factor analysis of high-dimensional and sparse matrices in Recommender Systems," *IEEE Transactions on Knowledge and Data Engineering*, vol. 35, no. 4, pp. 3897–3911, 2023.

[9] F. Bi, T. He, Y. Xie, and X. Luo, "Two-stream graph convolutional network-incorporated latent feature analysis," *IEEE Transactions on Services Computing*, pp. 1–15, 2023, DOI: 10.1109/TSC.2023.3241659.

[10] X. Luo, H. Liu, G. Gou, Y. Xia, and Q. Zhu, "A parallel matrix factorization based recommender by alternating stochastic gradient decent," *Engineering Applications of Artificial Intelligence*, vol. 25, no. 7, pp. 1403–1412, 2012.

[11] T. Maehara, K. Hayashi, and K.-ichi Kawarabayashi, "Expected tensor decomposition with stochastic gradient descent," *in Proc. of the AAAI Conference on Artificial Intelligence*, vol. 30, no. 1, 2016.

[12] X. Luo, M. C. Zhou, Y. N. Xia, and Q. S. Zhu, "An Efficient Non-negative Matrix-factorization-based Approach to Collaborative-filtering for Recommender Systems," *IEEE Transactions on Industrial Informatics*, vol. 10, no. 2, pp. 1273–1284, 2014.

[13] X. Chen, Z. He, and J. Wang, "Spatial-temporal traffic speed patterns discovery and incomplete data recovery via SVD-combined tensor decomposition," *Transportation Research Part C: Emerging Technologies*, vol. 86, pp. 59–77, 2018.

[14] H. Wu, X. Luo, M. C. Zhou, M. J. Rawa, K. Sedraoui, and A. Albeshri, "A PID-incorporated latent factorization of tensors approach to dynamically weighted directed network analysis," *IEEE/CAA Journal of Automatica Sinica*, vol. 9, no. 3, pp. 533–546, 2022.

[15] X. Luo, W. Qin, A. Dong, K. Sedraoui, and M. C. Zhou, "Efficient and high-quality recommendations via momentum-incorporated parallel stochastic gradient descent-based learning," *IEEE/CAA Journal of Automatica Sinica*, vol. 8, no. 2, pp. 402–411, 2021.

[16] D. Baidya, S. Dhopte, and M. Bhattacharjee, "Sensing System assisted novel PID Controller for efficient speed control of DC motors in Electric Vehicles," *IEEE Sensors Letters*, vol. 7, no. 1, pp. 1–4, 2023.

[17] L. Jin, X. Zheng, and X. Luo, "Neural Dynamics for distributed collaborative control of manipulators with time delays," *IEEE/CAA Journal of Automatica Sinica*, vol. 9, no. 5, pp. 854–863, 2022.

[18] D. Wu, X. Luo, M. Shang, Y. He, G. Wang, and M. C. Zhou, "A deep latent factor model for high-dimensional and sparse matrices in Recommender Systems," *IEEE Transactions on Systems, Man, and Cybernetics: Systems*, vol. 51, no. 7, pp. 4285–4296, 2021.

[19] H. Wu and X. Luo, "Instance-frequency-weighted regularized, nonnegative and adaptive latent factorization of tensors for dynamic QoS analysis," in *Proc. of the 2021 IEEE International Conference on Web Services (ICWS)*, Chicago, IL, USA , pp. 560-568, 2021.

[20] X. Luo, M. C. Zhou, Z. Wang, Y. Xia, and Q. Zhu, "An effective scheme for QoS estimation via alternating direction method-based matrix factorization," *IEEE Transactions on Services Computing*, vol. 12, no. 4, pp. 503–518, 2019.

[21] H. Wu, X. Luo, and M. C. Zhou, "Advancing non-negative latent factorization of tensors with diversified regularization schemes," *IEEE Transactions on Services Computing*, vol. 15, no. 3, pp. 1334-1344, 2020.

[22] W. Qin, H. Wu, Q. Lai, and C. Wang, "A Parallelized, Momentum-incorporated Stochastic Gradient Descent Scheme for Latent Factor Analysis on High-dimensional and Sparse Matrices from Recommender Systems," in *Proceedings of the 2019 IEEE International Conference on Systems, Man and Cybernetics (SMC)*, Melbourne, Australia, pp. 1744-1749, 2019.

[23] X. Luo, H. Wu, and Z. Li, "Neulft: A novel approach to nonlinear canonical polyadic decomposition on high-dimensional incomplete tensors," *IEEE Transactions on Knowledge and Data Engineering*, 2022, DOI: 10.1109/TKDE.2022.3176466.

[24] H. Li, P. Wu, N. Zeng, Y. Liu, and F. E. Alsaadi, "A survey on parameter identification, state estimation and data analytics for lateral flow immunoassay: From systems science perspective," *International Journal of Systems Science*, vol. 53, no. 16, pp. 3556–3576, 2022.

[25] X. Luo, M. C. Zhou, S. Li, and M. S. Shang, "An inherently nonnegative latent factor model for high-dimensional and sparse matrices from industrial applications," *IEEE Transactions on Industrial Informatics*, vol. 14, no. 5, pp. 2011–2022, 2018.

[26] Harshman, R. A. "Foundations of the PARAFAC procedure: Models and conditions for an" explanatory" multimodal factor analysis." (1970): 1-84.

[27] X. Luo, M. C. Zhou, H. Leung, Y. Xia, Q. Zhu, Z. You, and S. Li, "An incremental-and-static-combined scheme for matrix-factorization-based collaborative filtering," *IEEE Transactions on Automation Science and Engineering*, vol. 13, no. 1, pp. 333–343, 2016.

[28] S. Li, Y. Zhang, and L. Jin, "Kinematic control of redundant manipulators using neural networks," *IEEE Transactions on Neural Networks and Learning Systems*, vol. 28, no. 10, pp. 2243–2254, 2017.

[29] J. Chen, D. Ma, Y. Xu, and J. Chen, "Delay robustness of PID control of second-order systems: Pseudo concavity, exact delay margin, and performance tradeoff," *IEEE Transactions on Automatic Control*, vol. 67, no. 3, pp. 1194–1209, 2022.

[30] M. Whitby, L. Cardelli, M. Kwiatkowska, L. Laurenti, M. Tribastone, and M. Tschaikowski, "PID control of Biochemical Reaction Networks," *IEEE Transactions on Automatic Control*, vol. 67, no. 2, pp. 1023–1030, 2022.

[31] X. Luo, Y. Zhong, Z. Wang, and M. Li, "An alternating-direction-method of multipliers-incorporated approach to symmetric non-negative latent factor analysis," *IEEE Transactions on Neural Networks and Learning Systems*, pp. 1–15, 2021.

[32] S. Li, M. C. Zhou, X. Luo, and Z.-H. You, "Distributed winner-take-all in Dynamic Networks," *IEEE Transactions on Automatic Control*, vol. 62, no. 2, pp. 577–589, 2017.

[33] X. Luo, D. Wang, M. C. Zhou, and H. Yuan, "Latent factor-based recommenders relying on extended stochastic gradient descent algorithms," *IEEE Transactions on Systems, Man, and Cybernetics: Systems*, vol. 51, no. 2, pp. 916–926, 2021.

[34] Z. Cui, K. Henrickson, R. Ke, and Y. Wang, "Traffic graph convolutional recurrent neural network: A deep learning framework for network-scale traffic learning and forecasting," *IEEE Transactions on Intelligent Transportation Systems*, vol. 21, no. 11, pp. 4883–4894, 2020.

[35] W. Li, X. Luo, H. Yuan, and M. C. Zhou, "A momentum-accelerated Hessian-vector-based latent factor analysis model," *IEEE Transactions on Services Computing*, vol. 16, no. 2, pp. 830–844, 2023.

[36] L. Jin, L. Wei, and S. Li, "Gradient-based differential neural-solution to time-dependent nonlinear optimization," *IEEE Transactions on Automatic Control*, vol. 68, no. 1, pp. 620–627, 2023.

[37] X. Chen, Z. He, and L. Sun, "A bayesian tensor decomposition approach for spatiotemporal traffic data imputation," *Transportation Research Part C: Emerging Technologies*, vol. 98, pp. 73–84, 2019.

[38] X. Su, M. Zhang, Y. Liang, Z. Cai, L. Guo, and Z. Ding, "A tensor-based approach for the QoS evaluation in service-oriented environments," *IEEE Transactions on Network and Service Management*, vol. 18, no. 3, pp. 3843–3857, 2021.

[39] X. Luo, M. Chen, H. Wu, Z. Liu, H. Yuan, and M. Zhou, "Adjusting learning depth in nonnegative latent factorization of tensors for accurately modeling temporal patterns in dynamic QoS data," *IEEE Transactions on Automation Science and Engineering*, vol.18, no.4, pp. 2142-2155, 2021.

[40] H. Zhang, P. Chen, J. Zheng, J. Zhu, G. Yu, Y. Wang, and H. X. Liu, "Missing data detection and imputation for Urban ANPR system using an iterative tensor decomposition approach," *Transportation Research Part C: Emerging Technologies*, vol. 107, pp. 337–355, 2019.

[41] X. Luo, H. Wu, Z. Wang, J. Wang, and D. Meng, "A novel approach to large-scale dynamically weighted directed network representation," *IEEE Transactions on Pattern Analysis and Machine Intelligence*, vol. 44, no. 12, pp. 9756–9773, 2022.

[42] S. Li, M. Zhou and X. Luo, "Modified Primal-Dual Neural Networks for Motion Control of Redundant Manipulators with Dynamic Rejection of Harmonic Noises," in *IEEE Transactions on Neural Networks and Learning Systems*, vol. 29, no. 10, pp. 4791-4801, 2018.

[43] X. Luo, Y. Zhou, Z. Liu, L. Hu, and M. C. Zhou, "Generalized Nesterov's acceleration-incorporated, non-negative and Adaptive Latent Factor analysis," *IEEE Transactions on Services Computing*, vol. 15, no. 5, pp. 2809–2823, 2022.

[44] M. Liu, L. Chen, X. Du, L. Jin, and M. Shang, "Activated gradients for deep neural networks," *IEEE Transactions on Neural Networks and Learning Systems*, vol. 34, no. 4, pp. 2156–2168, 2023.

[45] X. Luo, Y. Yuan, S. Chen, N. Zeng, and Z. Wang, "Position-transitional particle swarm optimization-incorporated latent factor analysis," IEEE Transactions on Knowledge and Data Engineering, vol. 34, no. 8, pp. 3958–3970, 2022.

[46] J. Fang, Z. Wang, W. Liu, S. Lauria, N. Zeng, C. Prieto, F. Sikstrom, and X. Liu, "A new particle swarm optimization algorithm for outlier detection: Industrial data clustering in wire arc additive manufacturing," *IEEE Transactions on Automation Science and Engineering*, pp. 1–14, 2022, DOI:10.1109/TASE.2022.3230080.

[47] H. Wu, X. Luo, and M. C. Zhou, "Neural Latent Factorization of Tensors for Dynamically Weighted Directed Networks Analysis," in *Proceedings of the 2021 IEEE International Conference on Systems, Man and Cybernetics (SMC)*, Melbourne, Australia, pp. 3061-3066, 2021.